\documentclass[5pt]{article}

\usepackage[letterpaper]{geometry}
\usepackage{spconf,amsmath,epsfig}
\usepackage{enumitem}
\usepackage{subfigure}
\usepackage{multirow}

\usepackage{fancyhdr}
\begin{document}\sloppy

\def\x{{\mathbf x}}
\def\L{{\cal L}}

\title{Iterative Object and Part Transfer for Fine-Grained Recognition}
%
\name{Zhiqiang Shen, ~ Yu-Gang Jiang, ~ Dequan Wang ~ and ~ Xiangyang Xue}
\address{Shanghai Key Laboratory of Intelligent Information Processing,\\
		School of Computer Science, Fudan University  \\
        \tt\small \{zhiqiangshen13, ygj, dqwang12, xyxue\}@fudan.edu.cn \\ 
        }

%

\maketitle

\begin{abstract}
The aim of fine-grained recognition is to identify sub-ordinate categories in images like different species of birds. Existing works have confirmed that, in order to capture the subtle differences across the categories, automatic localization of objects and parts is critical. Most approaches for object and part localization relied on the bottom-up pipeline, where thousands of region proposals are generated and then filtered by pre-trained object/part models. This is computationally expensive and not scalable once the number of objects/parts becomes large. In this paper, we propose a nonparametric data-driven method for object and part localization. Given an unlabeled test image, our approach transfers annotations from a few similar images retrieved in the training set. In particular, we propose an iterative transfer strategy that gradually refine the predicted bounding boxes.  Based on the located objects and parts, deep convolutional features are extracted for recognition. We evaluate our approach on the widely-used CUB200-2011 dataset and a new and large dataset called Birdsnap. On both datasets, we achieve better results than many state-of-the-art approaches, including a few using oracle (manually annotated) bounding boxes in the test images. 
\end{abstract}
\section{Introduction}
\label{sec:intro}

Fine-grained recognition, also known as subcategory classification, has been actively studied in the past several years. In contrast to the traditional image category recognition, fine-grained recognition focuses on identifying sub-ordinate categories such as different species of birds. This rapidly growing subfield in image-based object recognition not only improves the performance of conventional methods, but also helps humans in specific domains, since some fine-grained categories can only be recognized by domain experts. 

\begin{figure}[t!]
	\begin{center}
		\includegraphics[width=0.6\linewidth]{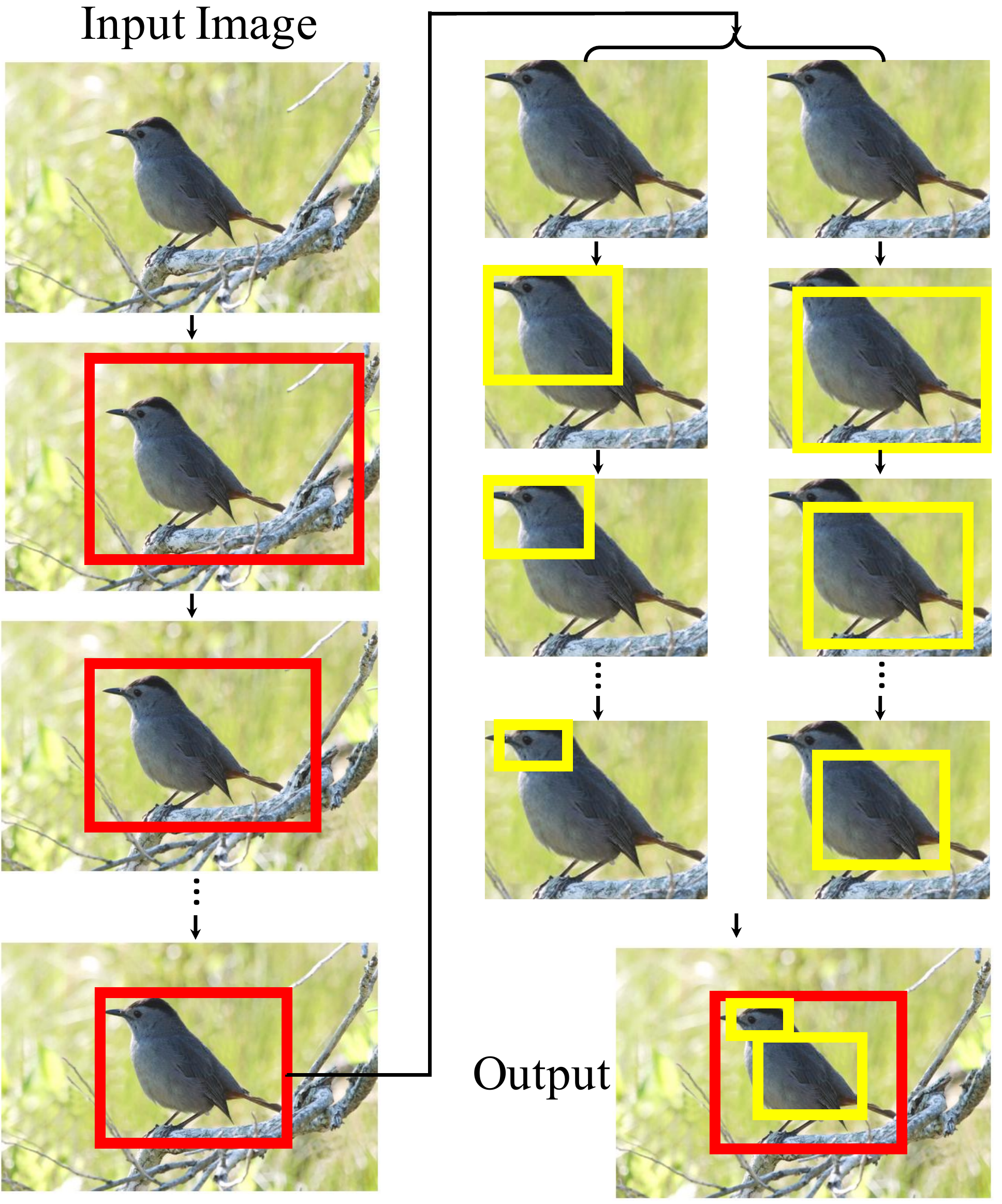}
	\end{center}
	\vspace{-0.2in}
	\caption{Illustration of the proposed approach, which gradually localizes objects (left) and their parts (right). The found objects and parts are used in fine-grained recognition.}
	\label{fig:short}
	\vspace{-0.1in}
\end{figure}

Traditional methods based on statistics of features calculated on the whole image~\cite{Dalal2005Histograms} are limited for fine-grained recognition, because there mainly exist subtle differences across the sub-ordinate categories. More effective solutions need to firstly localize the objects and their critical parts and then utilize features computed in the local regions for recognition~\cite{Zhang2014Part,berg-poof-cvpr2013,Branson2014Bird,Zhang2013Deformable}. In particular, the parts are often seen as discriminative regions, which are very important for capturing the subtle category differences. By focusing on the local regions, the effect of background clutter can also be largely alleviated, thus leading to outstanding recognition performance.

However, the large appearance variations that widely exist in the real-world make the task of object and part localization extremely challenging. The popular proposal-based localization approach like~\cite{felzenszwalb2010object} is not ideal as the filtering process of thousands of proposed candidate regions per image is expensive. In addition, it is difficult to train a robust ``filtering" model when the number of classes and parts becomes large. 

In this paper, we propose a novel approach that iteratively localizes objects and parts for fine-grained recognition. It follows the data-driven idea and is therefore model-free. The key idea is to ``transfer" location annotations from a few visually similar images retrieved in a large training dataset, where each image has bounding box annotations of both objects and important parts. One assumption in the data-driven approaches is that there exist a large amount of annotated data and, for most unseen test images, similar ones (in terms of both object and scene layout) can be found from the annotated training set so that the annotations can be reliably transferred to the unseen images. It is worth noting that this is not a very strong assumption in the big data era and similar pipelines have been successfully adopted in several related problems like image annotation~\cite{torralba200880} and human motion analysis~\cite{ren2005data}. 

In the data-driven localization process, we adopt an iteration based strategy to gradually focus on the target objects and the parts.  As shown in Figure \ref{fig:short}, our approach first locates a large bounding box of the bird object and then gradually adjusts the output towards a more precise localization boundary. The same method is also adopted to locate the parts. This iteration strategy is empirically found to be more effective than the existing method of one-step localization~\cite{goering2014nonparametric}.

The motivation behind our approach is very simple: when humans are given a visual scene, normally we obtain a gist of the scene first, and then gradually focus on specific objects. Object parts are probably only needed to be browsed or checked carefully if we want to understand the detailed properties of the object, \emph{e.g.}, the known clues to identify a particular species of bird. This biological visual perception procedure is simulated in the proposed approach for machine recognition of fine-grained categories.

\section{Related Work}
Fine-grained recognition has been extensively investigated recently. Most works used bird species categorization as the test case~\cite{huang2016part,berg-poof-cvpr2013,Zhang2014Part,Branson2014Bird,pu2014looks}, and some used leafs~\cite{Kumar2012Leafsnap}, flowers~\cite{cui2016fine} and dog breeds~\cite{Aditya2011}. Technically, one way to tackle the problem is to directly apply visual classification methods commonly used for standard object categorization. However, these approaches are incapable of capturing the subtle differences across the fine-grained categories. Thus, part-based approaches, which focus on extracting features in discriminative object parts, have become popular~\cite{Zhang2014Part,goering2014nonparametric}. One limitation of these approaches like~\cite{Zhang2014Part} is that they adopted a similar pipeline as~\cite{felzenszwalb2010object} for object/part detection, which relies on complex models that are difficult to be trained.

Based on the detected objects and parts, a few recent works focused on the extraction of more discriminative features~\cite{xiao2014application,Branson2014Bird}. For instance, a two-level attention model was proposed in~\cite{xiao2014application}. In addition, several researchers also explored the idea of human interaction based techniques~\cite{Wah2011Multiclass,Branson2011Strong}, which requires more manual inputs.

The main contribution of our work is the iterative data-driven approach for both object and part localization. A few existing works have also adopted the data-driven idea for localization, but used the one-step transfer process (without iteration) and many of them assumed that the object bounding boxes are given in the test images~\cite{goering2014nonparametric}. The key difference is that we utilize the iteration based strategy to gradually transfer object and part locations without requiring bounding box annotations at test time. A few researchers have investigated the idea of iterative learning in other problems like human pose estimation~\cite{carreira2015human}.

\section{The Proposed Approach}
We employ an iterative approach to process an image progressively from global to local regions. Our approach first locates the spatial areas of the objects and the object parts in the images. After that, we apply recognition models on the localized objects (and their parts) for category recognition. In the first step of localization, we adopt a data-driven scheme that reaches the goal by migrating information from similar images, where detailed category information is not needed. Specially, two levels of iterations are required in the localization step, which are elaborated in the following. 

\subsection{Localization}
\subsubsection{Object-level Transfer} 
\begin{figure}[t!]
	\begin{center}
		\includegraphics[width=0.9\linewidth]{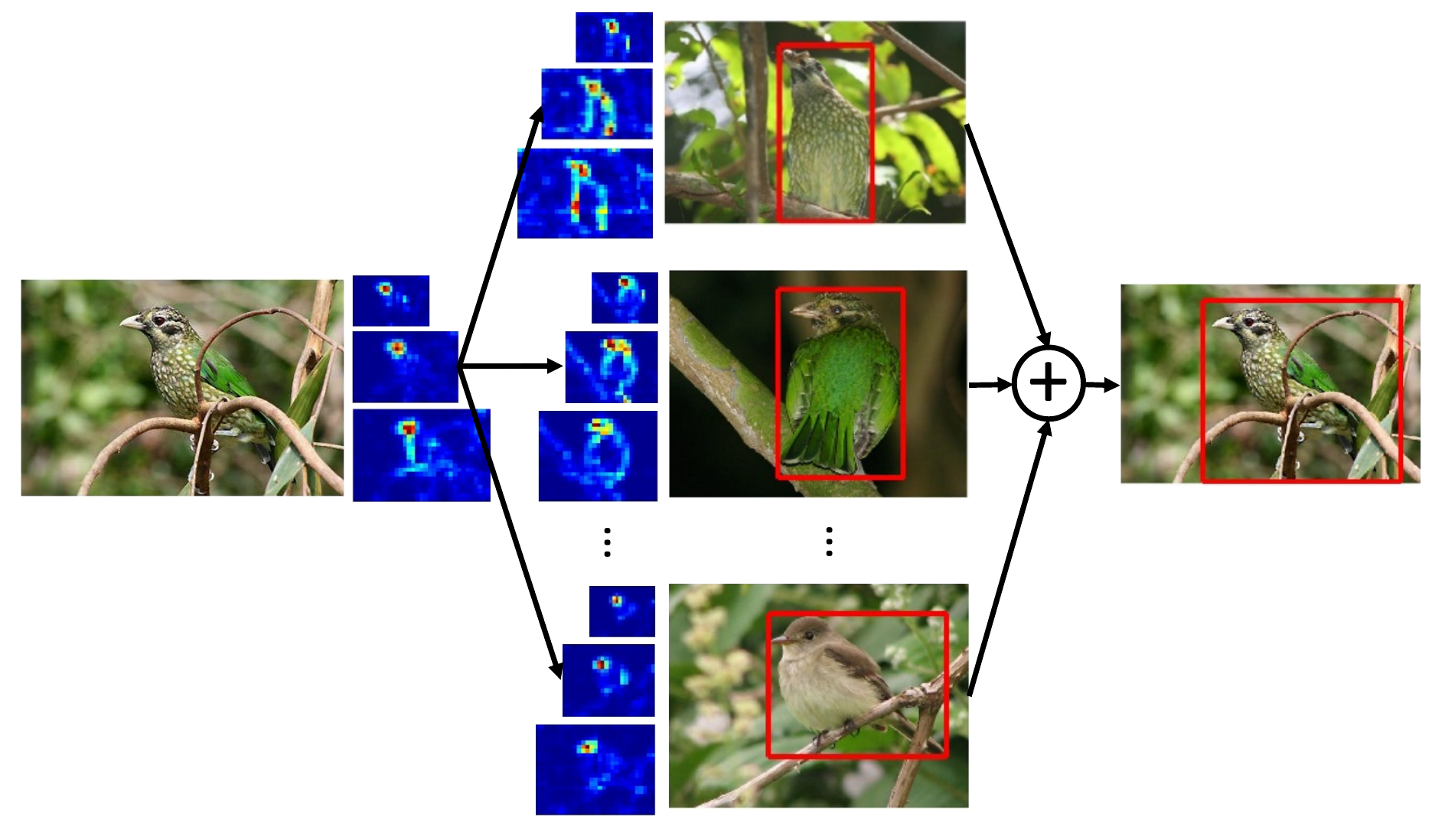}
	\end{center}
	\vspace{-0.25in}
	\caption{An iteration of the object-level transfer process. Bounding boxes of objects in similar training images (determined by matching CNN features) are transferred to an input image. The ``+" sign indicates bounding box fusion\&transfer. }
	\vspace{-0.1in}
	\label{fig:level1}
\end{figure}

We first use an iterative transfer scheme to locate the interested object in an input image. Figure~\ref{fig:level1} shows a single round of the transfer pipeline. The first step is to extract image features at multiple scales. For this, we adopt the popular convolutional neural networks (CNNs) using a publicly available network model called VGGNet \cite{simonyan2014very}. We follow the recent work of~\cite{he2014spatial} to extract the CNN features, where an spatial pyramid pooling (SPP) layer is added on top of the last convolutional layer, which pools features and generates fixed-length outputs.

Based on the features computed from the input image, we retrieve a small set of \emph{nearest neighbors} in the training dataset, where the images are labeled with both object and part locations. Next, the location annotations from the similar training images are transferred to the input image. Since the images and the objects are of different sizes, we propose a simple bounding box fusion method so that the location annotations from multiple training images can be combined to produce the bounding box for the input image. 

Specifically, the bounding box fusion process is executed by mapping all the images into a common space and then merging the boxes. Given an input image $I$ with its corresponding size information $S$, we have a candidate set of $M$ images, denoted by ${\left\{ {\;{b^i},\;{s^i}} \right\}_{i = 1,...,M}}$ where ${b^i}$ and ${s^i}$ are  the bounding box annotations and the sizes of candidate images.  All the images are resized into a uniform size $s_{uni}$ with the bounding box locations updated according to the new size, denoted by ${\left\{ {\;{b^i_{uni}},\;{s_{uni}}} \right\}_{i = 1,...,M}}$. We then take the union of the bounding boxes $\cup_i b^i_{uni}$ as the fused bounding box of the input image. Finally, this fused box can be mapped back according to the original size of the input image as the output of this iteration. Notice that union is used as we found it more effective than average or intersection fusion, because it maximizes the likelihood of containing the entire object. 

\begin{figure}[t]
	\begin{center}
		\subfigure[Cropped boxes before adjustment.]{\includegraphics[height=0.05\textwidth]{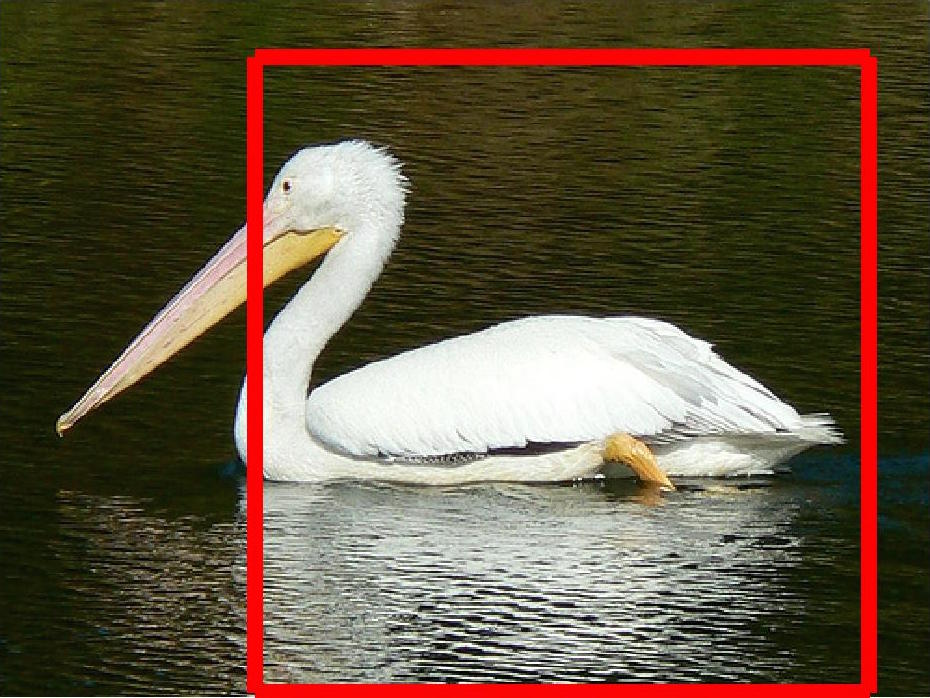}
			\includegraphics[height=0.05\textwidth]{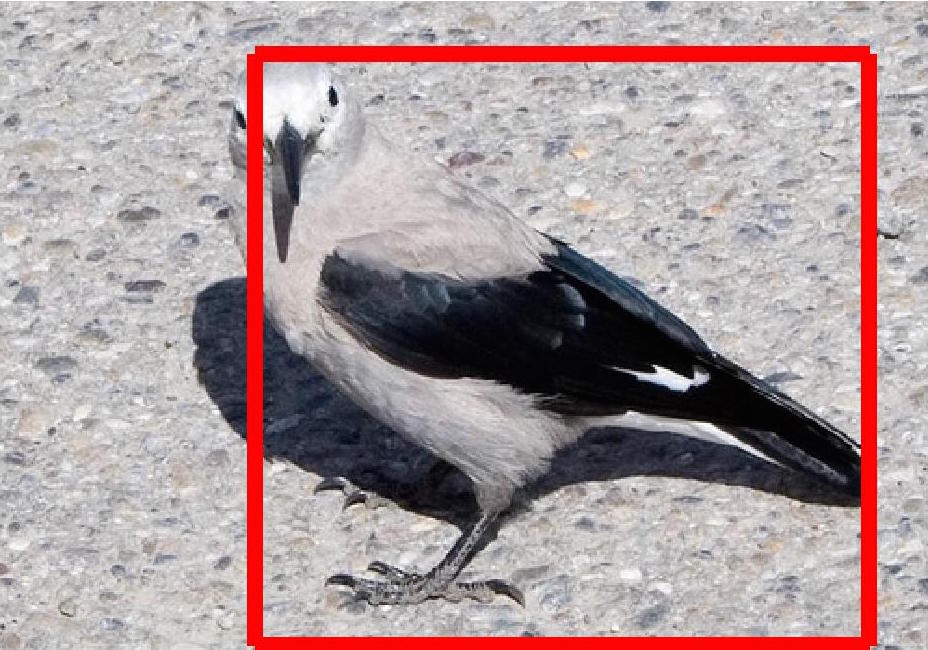}
			\includegraphics[height=0.05\textwidth]{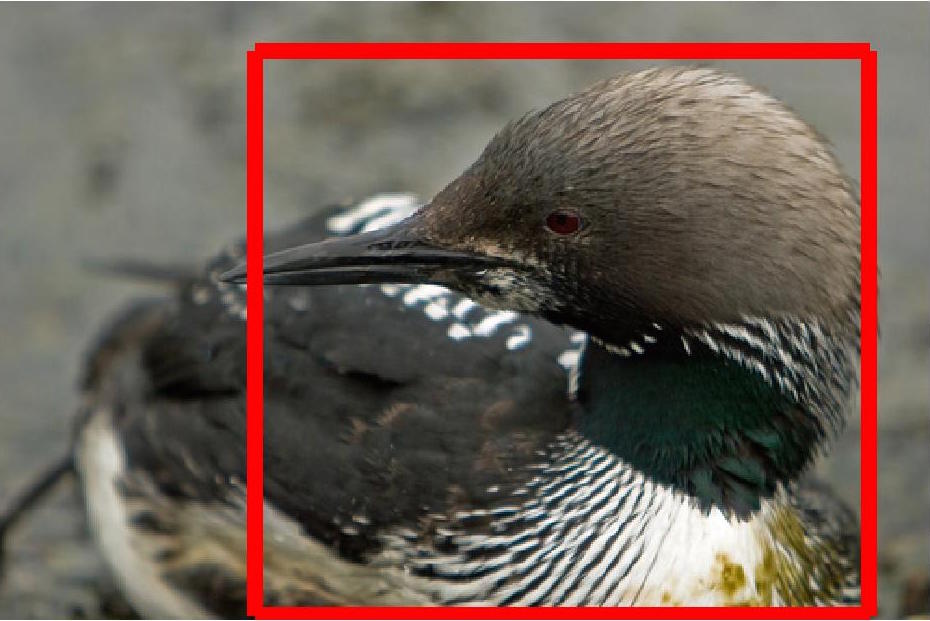}
			\includegraphics[height=0.05\textwidth]{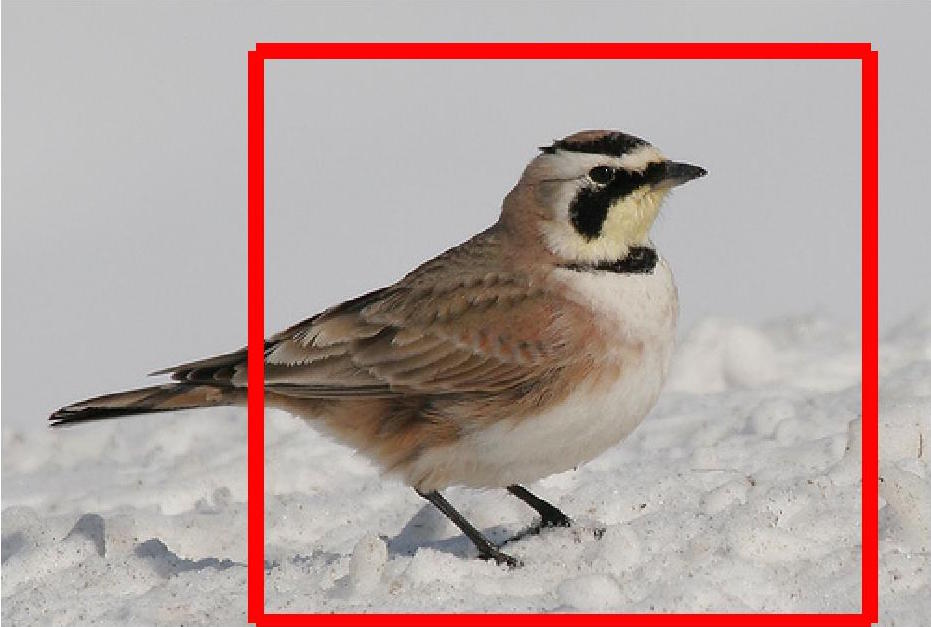}
			\includegraphics[height=0.05\textwidth]{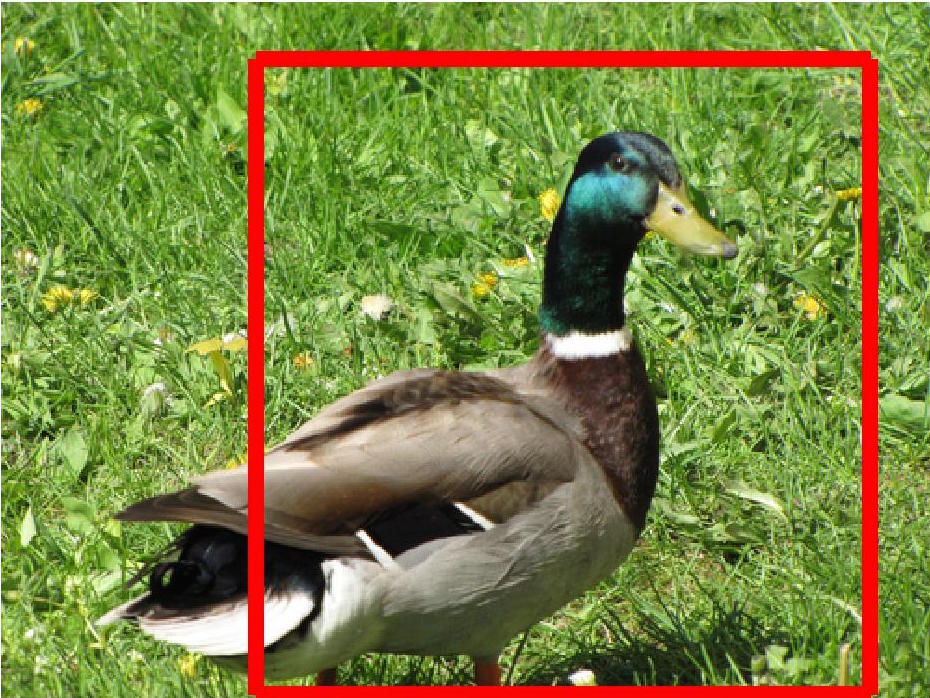}
			\includegraphics[height=0.05\textwidth]{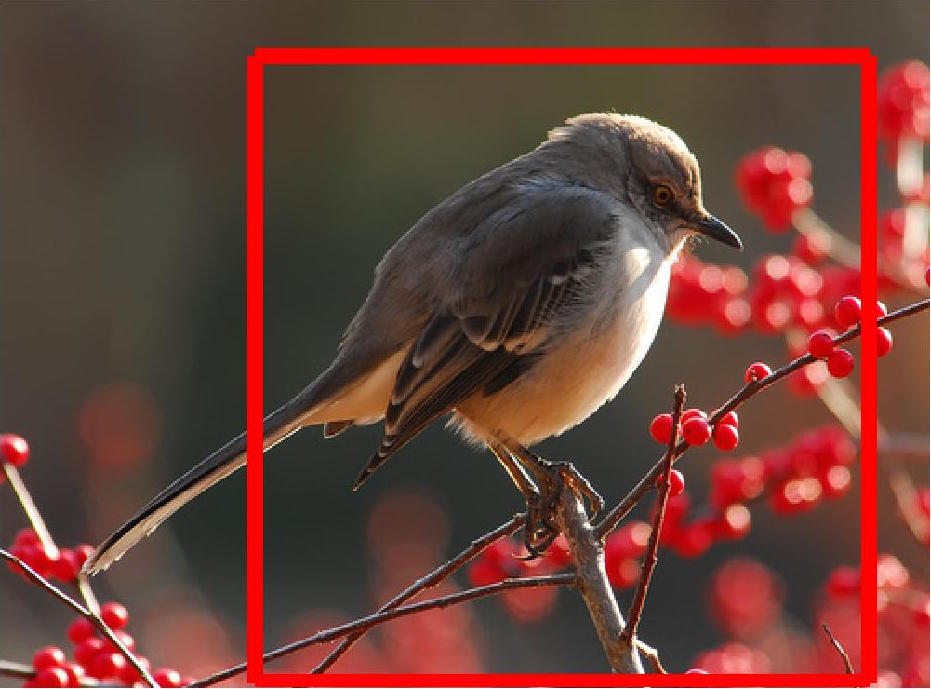}}
		
		\subfigure[Cropped boxes after adjustment.]{\includegraphics[height=0.05\textwidth]{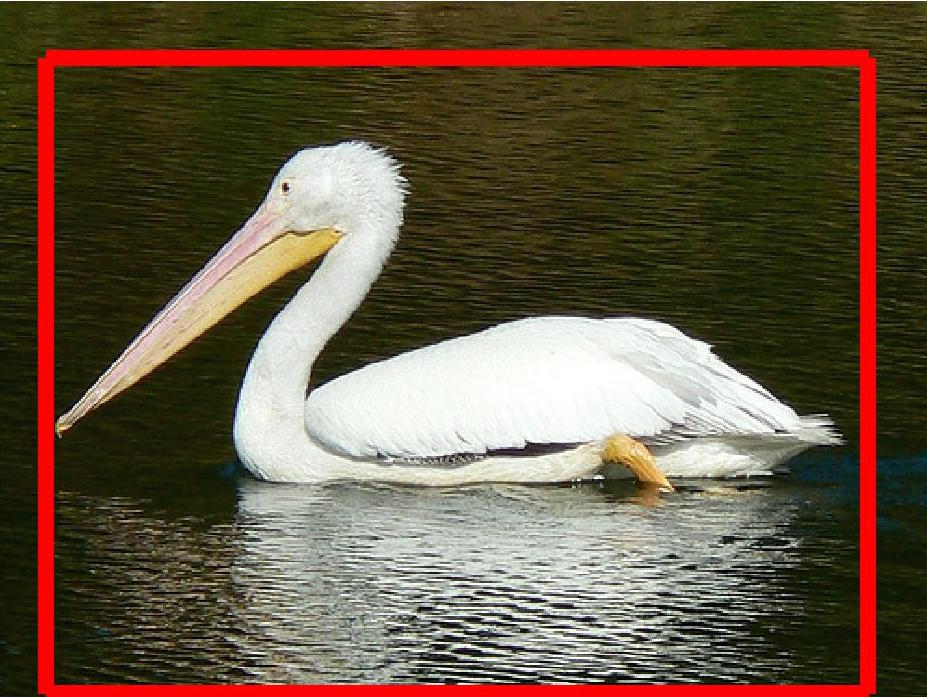}
			\includegraphics[height=0.05\textwidth]{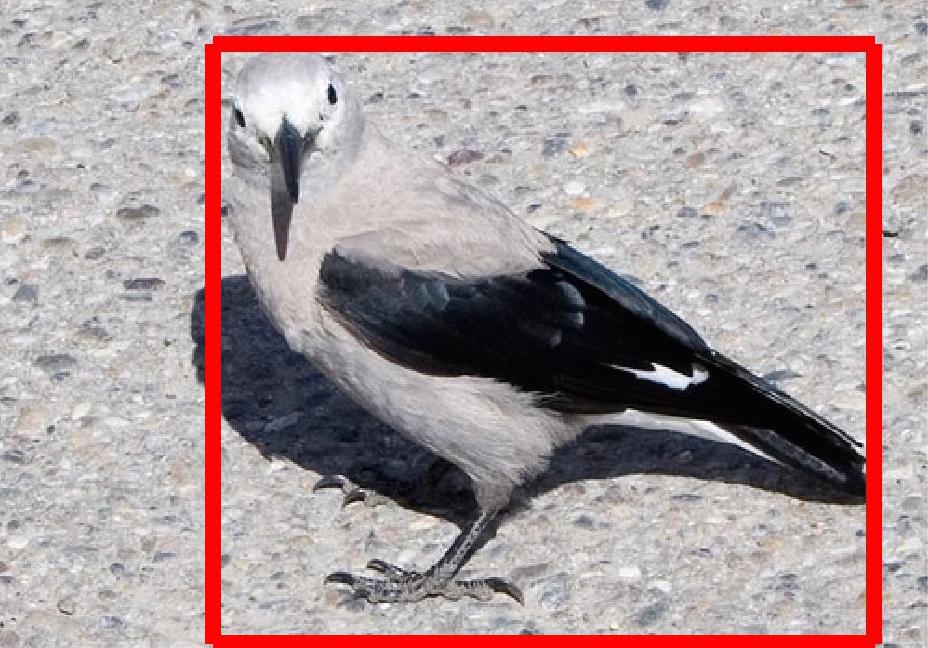}
			\includegraphics[height=0.05\textwidth]{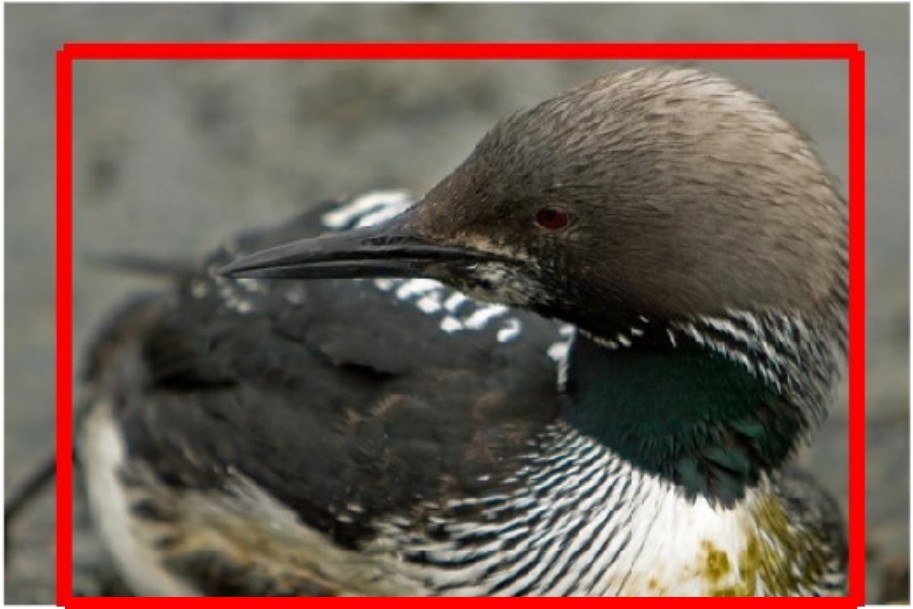}
			\includegraphics[height=0.05\textwidth]{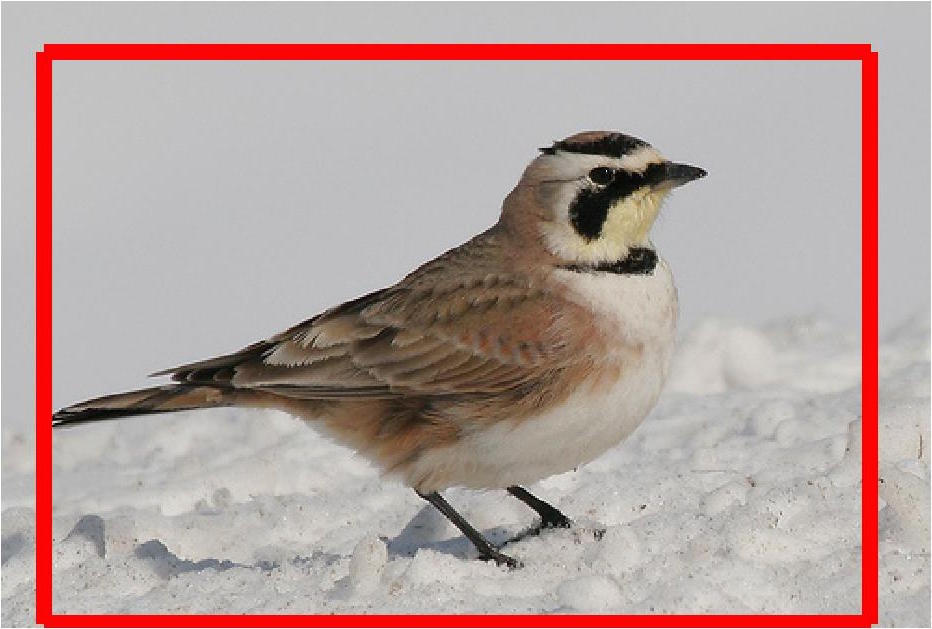}
			\includegraphics[height=0.05\textwidth]{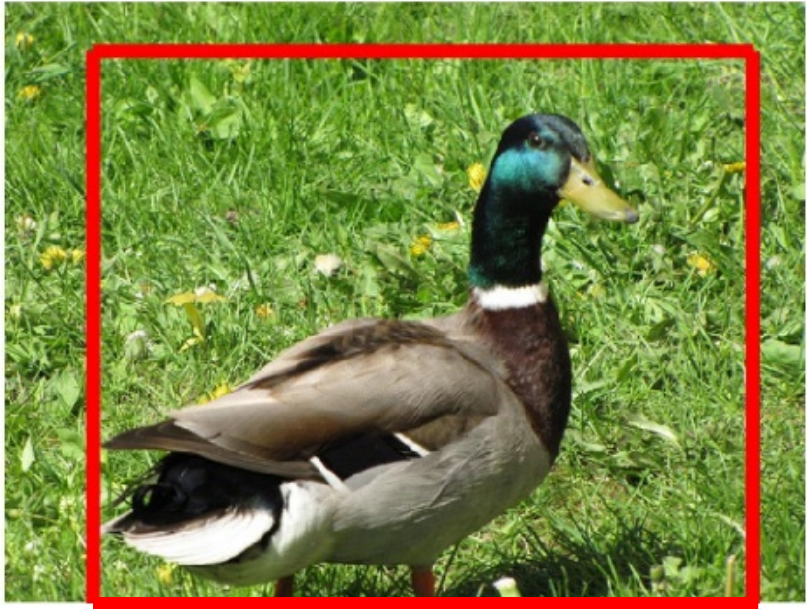}
			\includegraphics[height=0.05\textwidth]{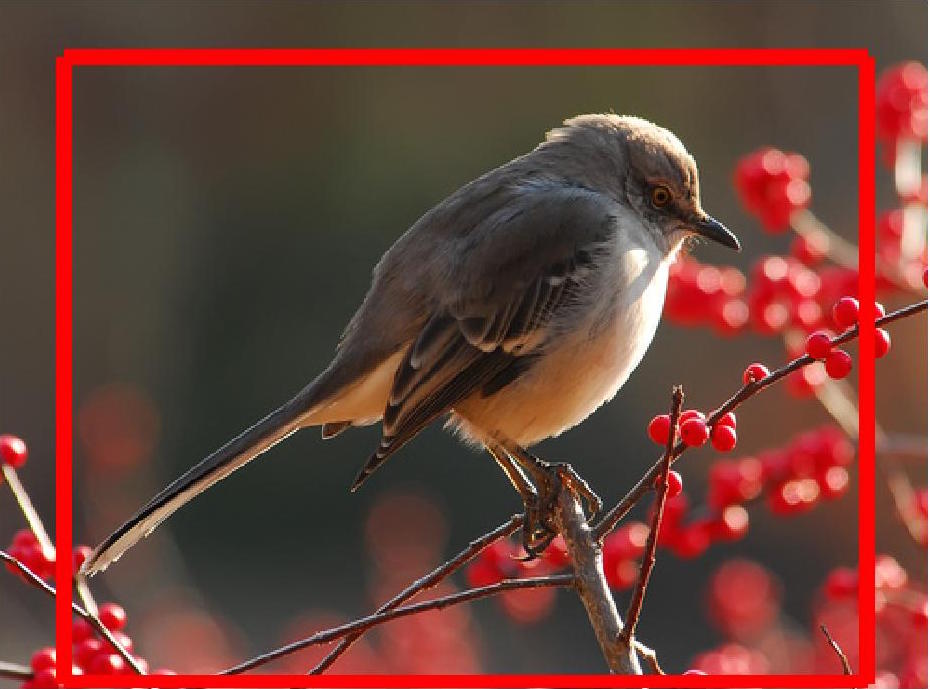}}
		
	\end{center}
	\vspace{-0.22in}
	\caption{Adjusting cropped boxes in training images (to ensure the entire objects are preserved) for object-level transfer. See texts for more explanations.}
	\label{fig:fusion}
	\vspace{-0.15in}
\end{figure}

After receiving the bounding box from the first iteration, we update the input image by cropping out only the object areas, with which we proceed to perform the next iteration to generate a more precise bounding box. Before performing the next iteration, we also crop all the training images so that they can be matched more accurately with the input image. This is done by treating each training image as an input image, and using the rest to transfer the bounding boxes. In order to ensure that all the cropped training images contain the entire objects, we adjust the cropped area using the bounding box annotations, as visualized in Figure~\ref{fig:fusion}.

There are multiple ways to terminate this iteration process. One way is to stop when the bounding box does not change significantly across different iterations. As the bounding boxes are eventually used for recognition in our problem, we adopt a different strategy when the prediction score from a raw classifier trained on entire images (not the detected bounding boxes) is higher than a pre-defined threshold. This is easy to implement and was found slightly better.

\subsubsection{Part-level Transfer} 
The object-level bounding boxes are not sufficient for fine-grained recognition as the differences across some categories may only lie in very small object parts. Harnessing features computed on such parts will be very helpful, which has been validated by several previous studies~\cite{berg-poof-cvpr2013,R.2011Birdlets,Zhang2014Part,Zhang2013Deformable,Liu2012Dog}. In this work, we execute a similar iterative process like the object-level transfer to locate critical object parts.

\begin{figure}[t]
	\begin{center}
		\includegraphics[width=0.9\linewidth]{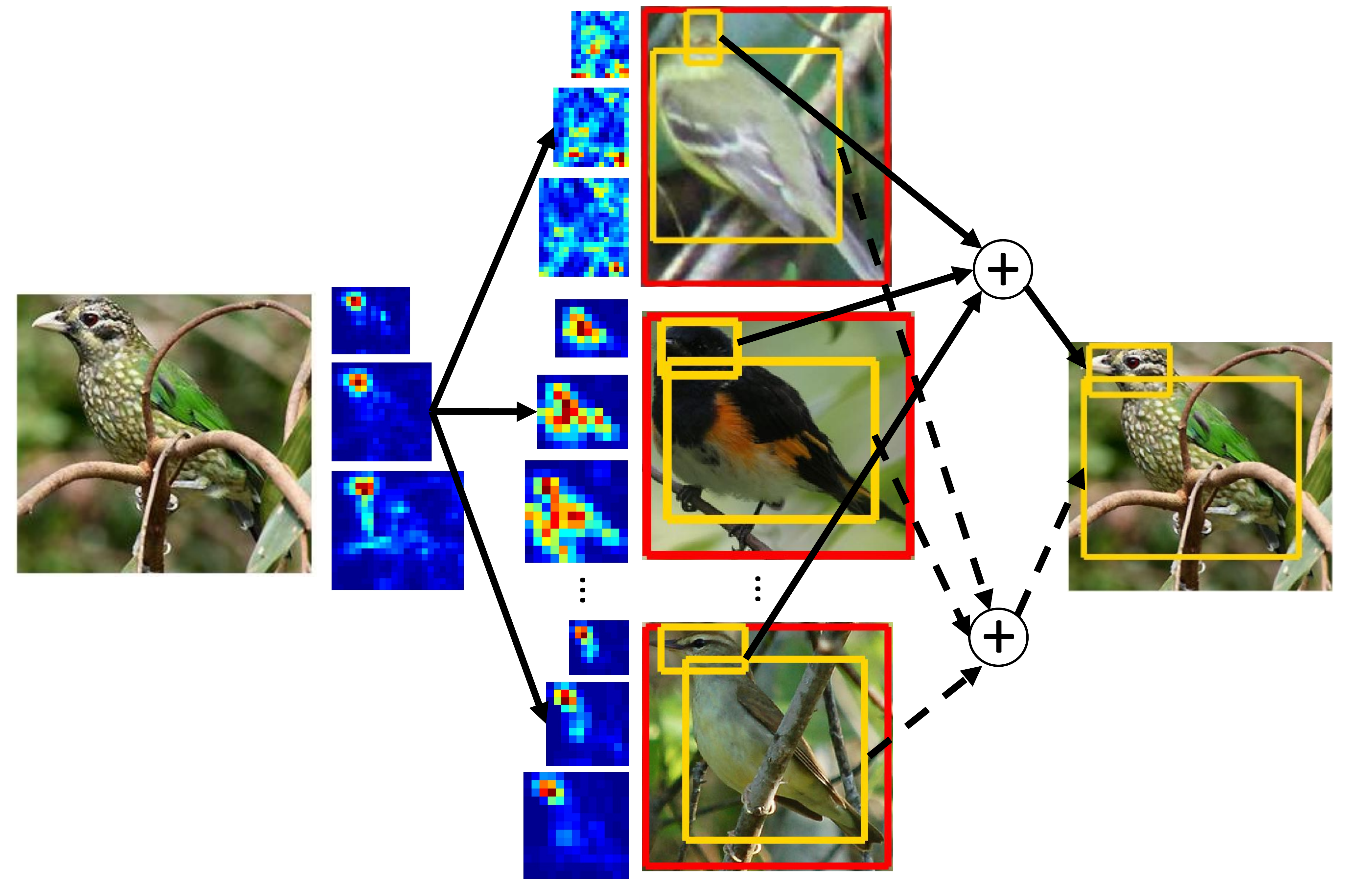}
	\end{center}
	\vspace{-0.2in}
	\caption{An iteration of the part-level transfer process. Bounding boxes of parts in similar training images (determined by matching CNN features) are transferred to an input object.}
	\vspace{-0.1in}
	\label{fig:level2}
\end{figure}

Our part-level transfer pipeline is shown in Figure~\ref{fig:level2}. In this pipeline, we take the localized object as input and compare against the objects in the training set. Similar images are found based on matching the same CNN features. The part-level bounding boxes are fused in the same way as we fuse the object-level bounding boxes. This process can be iteratively executed to achieve a good localization of parts.

We underline that our localization approach is quite different from the proposal-based methods \cite{Zhang2014Part}, which extract thousands of candidate boxes in one image and filters all of them to pick the most possible object bounding box(es). Our method relies on a purely data-driven method, which is much easier to be implemented and, as will be shown later, performs even better.

\subsubsection{Bounding Box Refinement with Regression}\label{reg}
\begin{figure}[t]
	\begin{center}
		\includegraphics[width=0.97\linewidth]{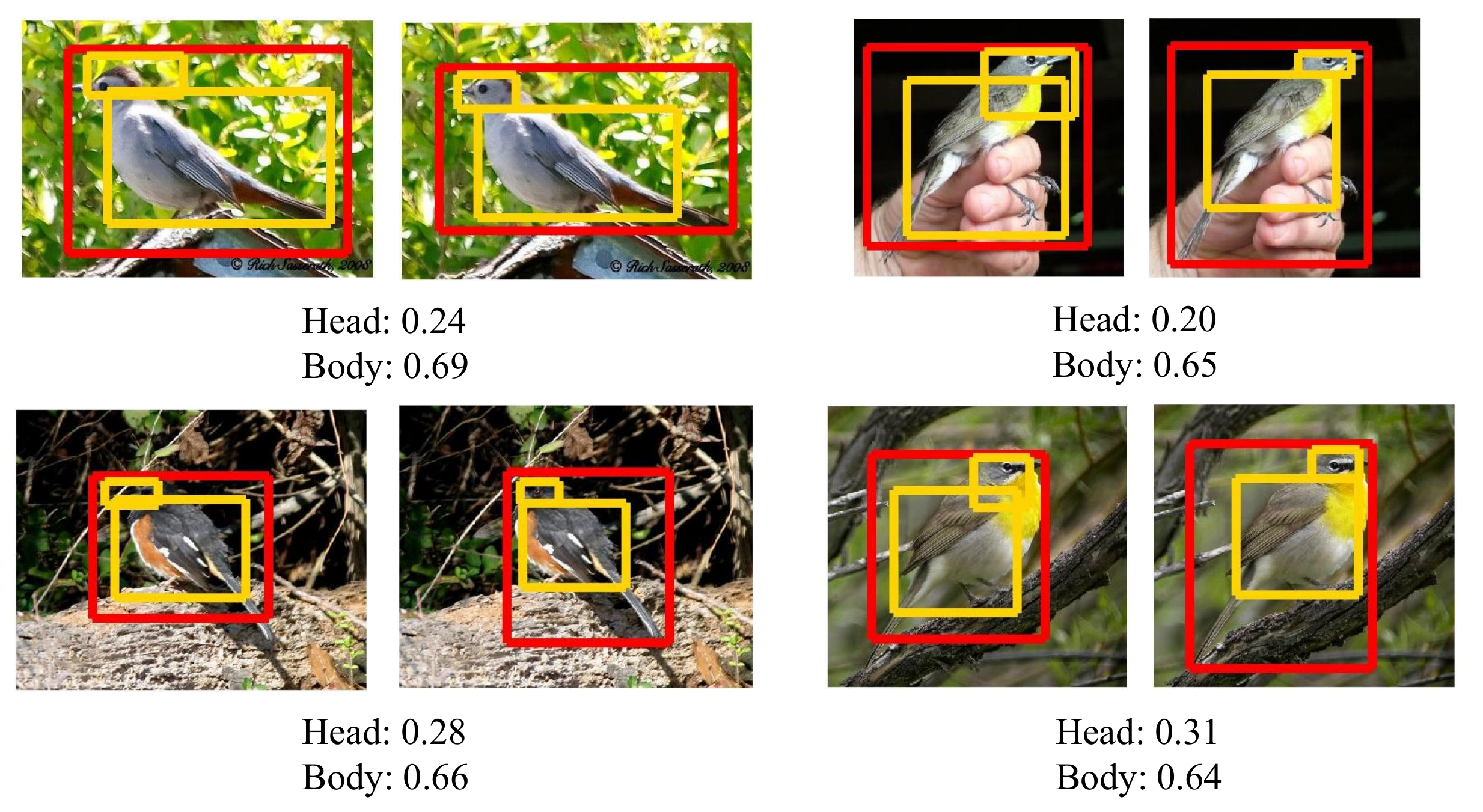}
	\end{center}
	\vspace{-0.2in}
	\caption{Four examples of detected boxes (left) and ground-truth boxes (right). The Intersection-over-Union values of the two part-level boxes are shown below the images.}
	\label{fig:overlap}
	\vspace{-0.1in}
\end{figure}

The bounding boxes obtained by the proposed two-level iterative process are good but there is still room for improvement. A popular measure to evaluate the quality of object/part localization is Intersection-over-Union, which computes the percentage of the overlapped region between the detected box and the ground-truth box over the union of the two boxes. Figure~\ref{fig:overlap} gives a few examples, where we see that the measure is pretty low for small boxes like the head of the birds. 

We use a simple bounding box regression method to mitigate the deviation. Based on the object and part bounding boxes obtained by the iterative process, we predict refined bounding boxes using a class-specific bounding box regressor. This is similar to the method used in previous works like R-CNN~\cite{Girshick2013Rich} and deformable part models~\cite{felzenszwalb2010object}.

More formally, our goal is to learn a transformation that maps the predicted box to the corresponding ground-truth box.  Suppose there are $N$ training pairs $\{(T^i,G^i)\}_{i=1,...,N}$, where $T=(T_x,T_y,T_w,T_h)$ denotes the coordinates of the predicted boxes together with width and height and $G=(G_x,G_y,G_w,G_h)$ denotes the ground-truth boxes.

Following~\cite{Girshick2013Rich}, the transformation is parameterized using four functions $f_x(T)$, $f_y(T)$, $f_w(T)$ and $f_h(T)$, where the first two refer to the scale invariant translation of the box coordinates (upper-left corner) and the last two indicate the log-scale translations of the width and height of the box. Once these functions are learned, the refined box (predicted ground-truth) can be obtained by: ${{\hat G}_x} = {T_w}{f_x}(T) + {T_x}$, ${{\hat G}_y} = {T_h}{f_y}(T) + {T_y}$, ${{\hat G}_w} = {T_w}{e^{{f_w}(T)}}$, ${{\hat G}_h} = {T_h}{e^{{f_h}(T)}}$.

Each function $f(\cdot)$ is modeled in linear form with the CNN features as input: ${f_*}(T) = {w_*}{\phi}(T)$, where $*$ indicates one of $x,y,h,w$ and $\phi$ is the CNN feature. $w_*$ is the vector of parameters, which are learned by optimizing the following objective function:
\begin{equation}
{w_*} = \mathop {\arg \min }\limits_{{{\hat w}_*}} {\sum\limits_i^N {{{(Y_*^i - {{\hat w}_*}{\phi}({T^i}))}^2} + \lambda \left\| {{{\hat w}_*}} \right\|^2}}, 
\end{equation}
where the regression target $Y_*$ is defined as $Y_*=(G_*-T_*)/T_*$ if $*$ is $x$ or $y$ and $Y_*=log(G_*/T_*)$ if $*$ is $w$ or $h$. 

\subsection{Recognition} 
After the iterative transfer and the bounding box refinement, we arrive at a set of object and part bounding boxes for each input image containing an interested object\footnote{In practice, an image without a target interested object may be excluded at the localization stage if it has small matching similarity scores with the training images.}. To recognize the specific type or class of the object, we also adopt the CNN features computed in each object/part bounding box. The VGGNet model \cite{simonyan2014very} is adopted with parameters fine-tuned using the image patches in the bounding boxes. Features extracted by the fine-tuned CNN model from different boxes are concatenated to train one-vs-all linear SVM classifiers for final prediction. Notice that this simple feature concatenation based recognition method has been adopted by several previous works \cite{goering2014nonparametric,Zhang2014Part}. Advanced fusion methods that automatically learn the weights of each feature \cite{mm14:videoclassification} may lead to better performance.


\section{Experiments}
\subsection{Dataset and Evaluation}
In this section, we evaluate our approach on the widely-used fine-grained recognition benchmark CUB200-2011~\cite{Wah2011The}. We also report results on a new, large dataset called Birdsnap~\cite{berg-birdsnap-cvpr2014}.
\begin{itemize}	
	\item CUB200-2011 (a.k.a. Caltech-UCSD Bird) dataset contains 11,788 images of 200 bird species. Each image in CUB200-2011 is annotated with bounding boxes of both object (bird) and parts. We adopt two part boxes in the experiments: head and body, following the protocol of~\cite{Zhang2014Part}.
	
	\item Birdsnap is a much larger dataset with 49,829 images spanning 500 species of North American birds. Each image has detailed location annotations and additional attribute labels such as male, female, immature, {\em etc}. In this work, we only adopt the location annotations.
\end{itemize}

For both datasets, localization accuracy is measured by the percentage of correctly localized parts (PCP). A detected part is considered as a correct hit only when its Intersection-over-Union value with the ground-truth is larger than a threshold. Object-level localization results are not discussed as localizing parts is a more difficult task, and once parts are correctly localized, object localization is most likely to be correct. 

For the final recognition results, we also use accuracy as the performance measure, which is the percentage of samples with correctly recognized bird species. 

\subsection{Results on CUB200-2011}
We first report and discuss results evaluating in isolation the ability of our approach to accurately localize parts, which are summarized in Table~\ref{M_score} and Table~\ref{part-location}. After that, we present recognition results using different kinds of inputs in Table~\ref{part-result} and compare with the state of the arts in Table~\ref{CUB-2011results}.

\subsubsection{Part Localization} 
In Table~\ref{M_score}, we summarize the results using different {\em M}, \emph{i.e.}, the number of nearest training images used for bounding box transfer. For all the evaluated overlapping thresholds, $M=2$ seems a good option. Using a single most similar image in the training set and copying its bounding boxes is not precise enough, while using too many training images may involve noise from the less similar ones. The head part localization results are lower than that of body as heads are smaller and a small location shift away from the ground-truth may affect significantly on the overlapping ratio (see Figure \ref{fig:overlap}). 

\begin{figure}[t!]
	\begin{center}
		\begin{minipage}{1\textwidth}	
			\includegraphics[height=0.102\linewidth]{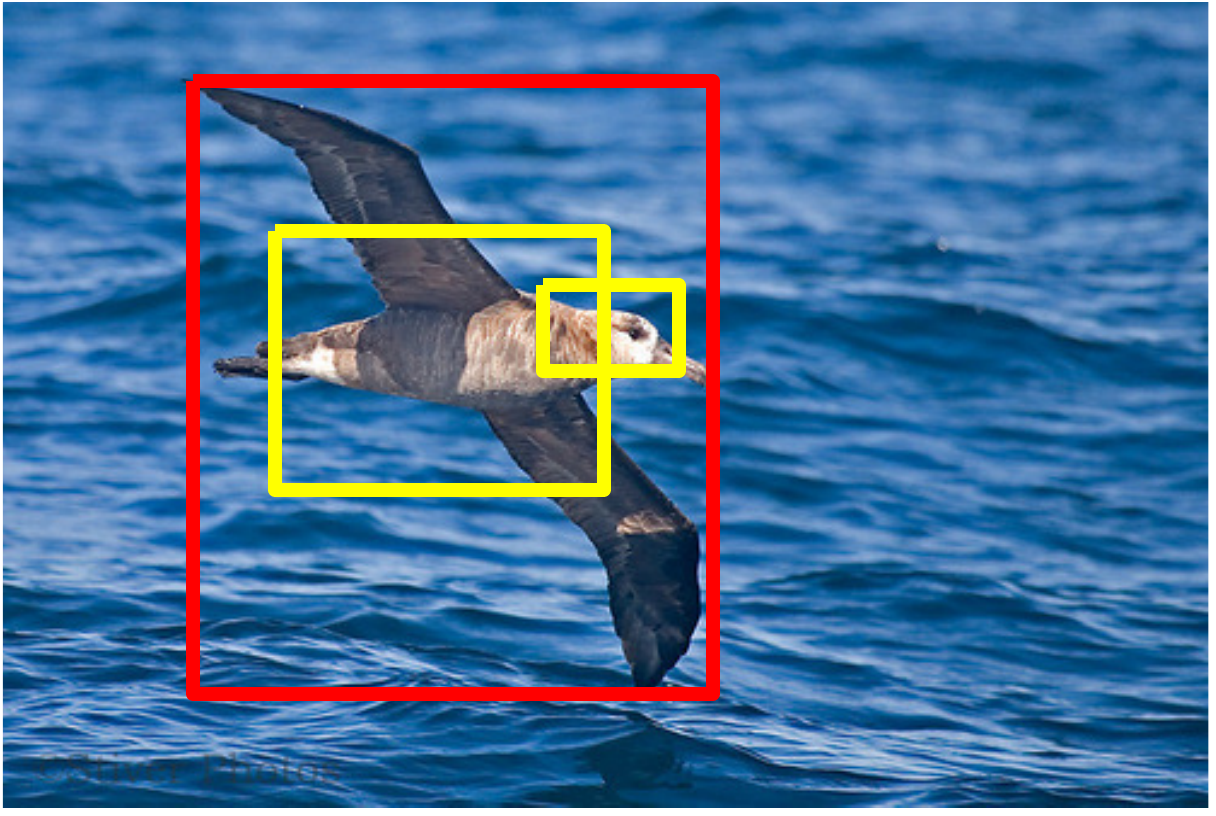}
			\includegraphics[height=0.102\linewidth]{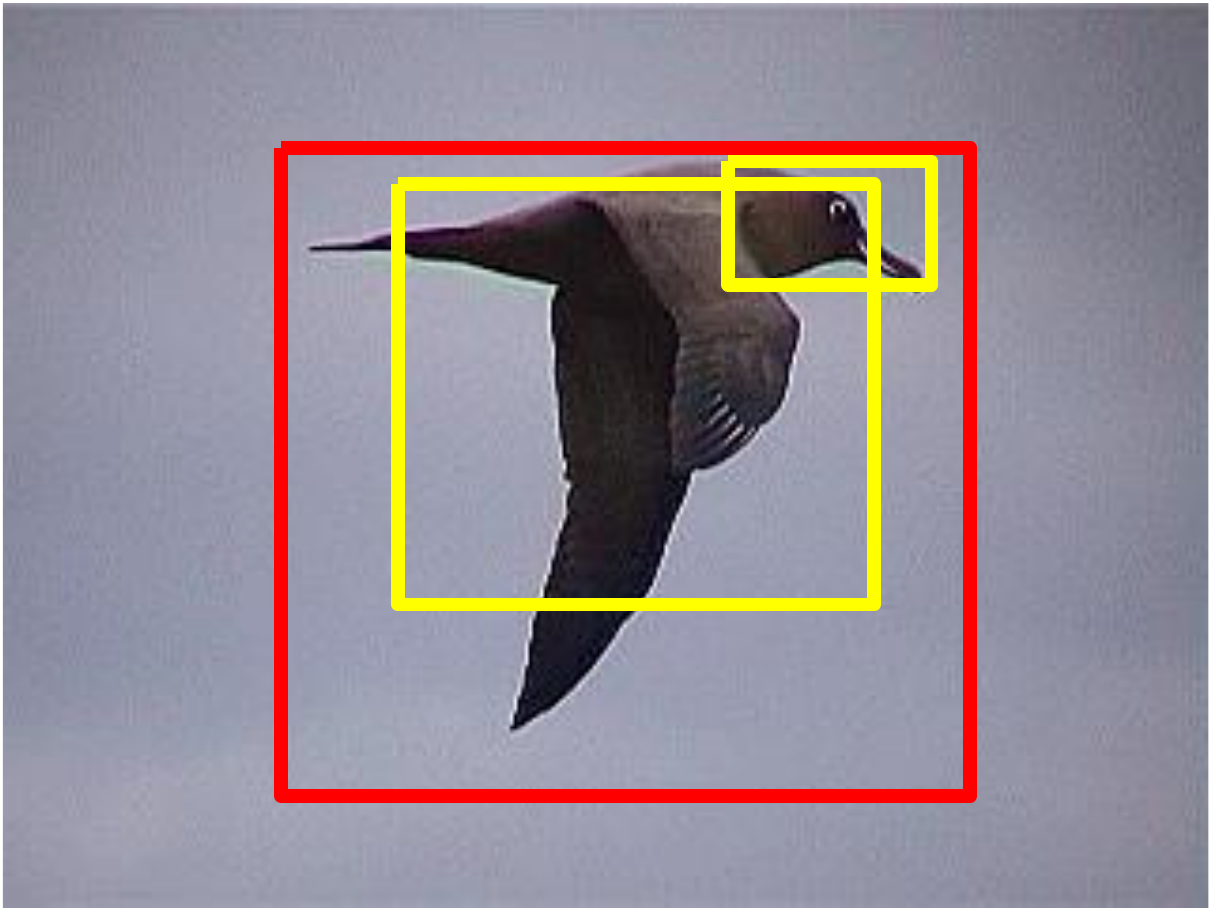}
			\includegraphics[height=0.102\linewidth]{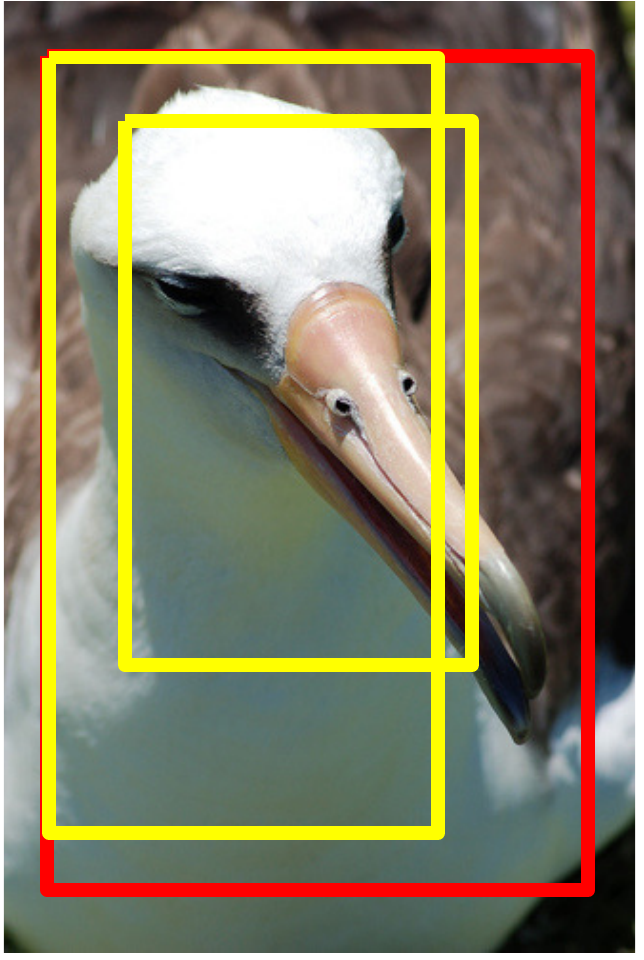}
			\includegraphics[height=0.102\linewidth]{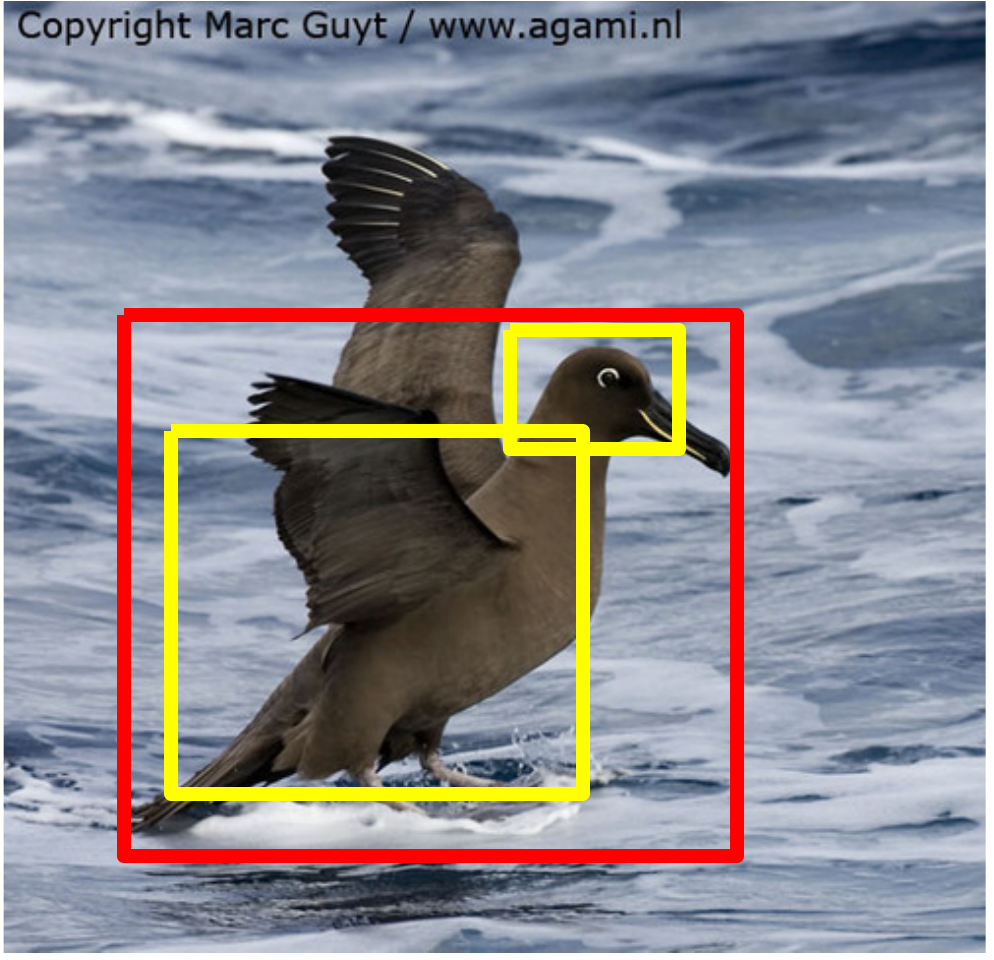}
			
			\includegraphics[height=0.098\linewidth]{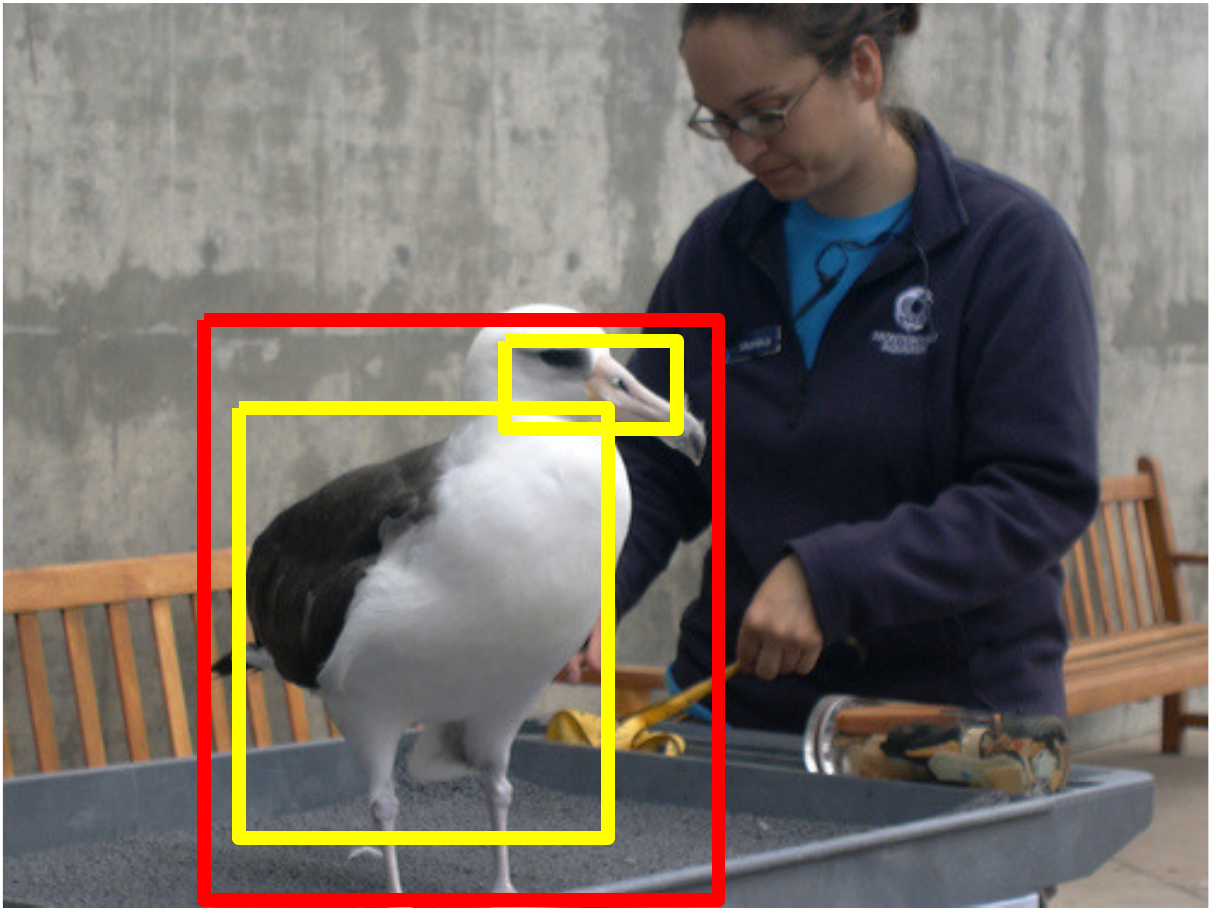}
			\includegraphics[height=0.098\linewidth]{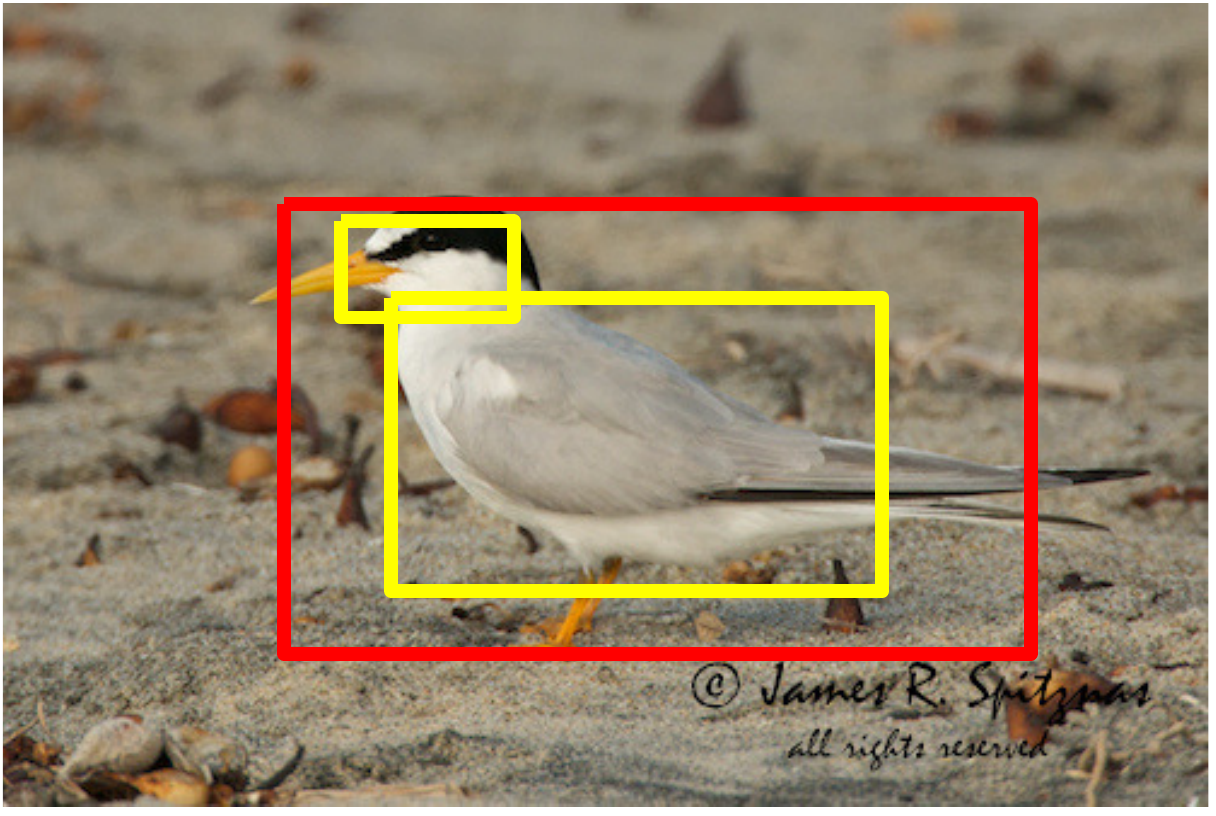}
			\includegraphics[height=0.098\linewidth]{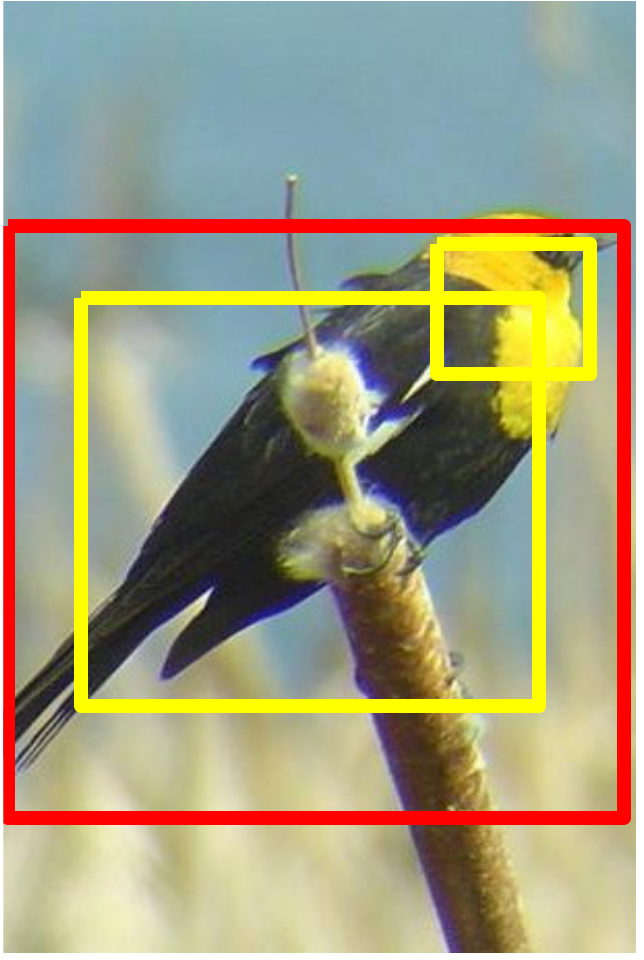}
			\includegraphics[height=0.098\linewidth]{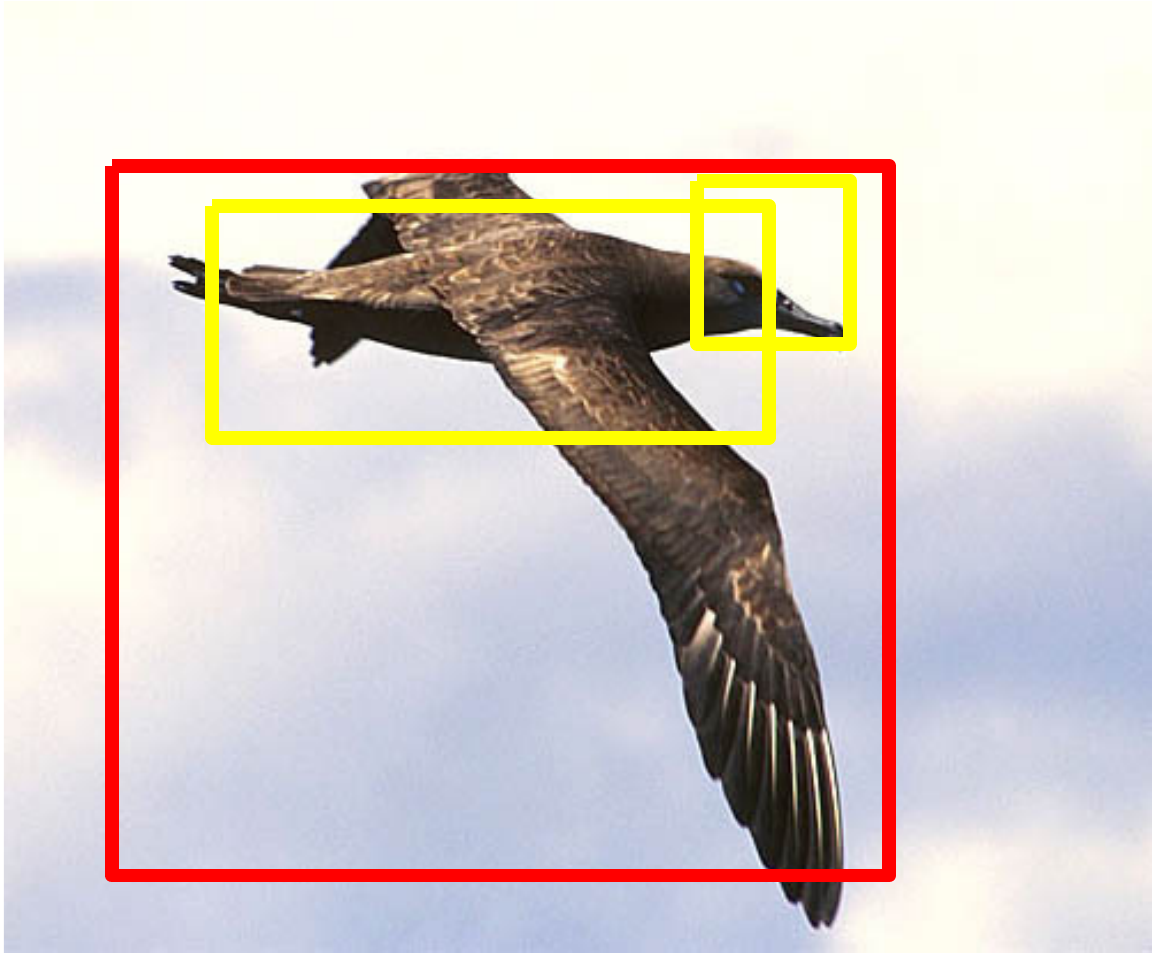}
		\end{minipage}
	\end{center}
	\vspace{-0.2in}
	\caption{Examples of localization results using our proposed approach. The red bounding box indicates the detected object and the yellow ones indicate the head and body parts.}
	\label{fig:localizationExamples}
	\vspace{-0.2in}
\end{figure}

We also evaluate the results of part localization by assuming the object-level bounding box is given. Results are reported in Table~\ref{part-location}, together with the results of two compared representative approaches: DPM~\cite{Azizpour2012Object} and R-CNNs~\cite{Zhang2014Part}. We see that the results of both compared approaches are significantly better when the object-level bounding boxes are given. In contrast, our approach holds the very appealing advantage of not requiring the oracle object-level boxes as inputs----the performance of not knowing the oracle object bounding boxes is similar under most settings.  Figure~\ref{fig:localizationExamples} shows several examples of our localization results. 

Compared with the two alternative approaches, we obtain significantly better results for the body part and lower accuracy for the head part (using the same overlapping threshold $\geq$0.5). The reason of our low performance of head detection is that we take the ``union" of the training bounding boxes in the iterative transfer process, which normally produces larger boxes. This is fine for large parts like body, but for small parts, as discussed earlier, the Intersection-over-Union values of the predicted boxes are affected much more significantly. Note that the slightly larger bounding boxes from our approach turn out to be better in the recognition stage (see comparison of recognition results with the same approach in Table~\ref{CUB-2011results}), which may be due to the fact that the ground-truth annotations are not very accurate and tend to be smaller than the real object parts in many cases.

\begin{table}[t!]
	\centering
	\resizebox{0.9\hsize}{!}{%
		\begin{tabular}{|c|c|c|c|c|c|c|}
			\hline
			&             \multicolumn{3}{c|}{Head}              &              \multicolumn{3}{c|}{Body}              \\ \hline
			& $\geq0.5$ & $\geq0.4$ & $\geq0.3$ & $\geq0.5$ & $\geq0.4$ & $\geq0.3$ \\ \hline
			M=1 &       43.9       &       63.7       &       77.4       &       82.5       &       89.8       &       94.6       \\ \hline
			M=2 &  \textbf{50.1}   &  \textbf{68.0}   &  \textbf{80.5}   &  \textbf{85.9}   &  \textbf{92.1}   &  \textbf{96.0}   \\ \hline
			M=3 &       46.0       &       64.2       &       78.0       &       85.5       &       91.6       &       95.7       \\ \hline
			M=4 &       39.7       &       59.5       &       75.5       &       84.4       &       91.1       &       95.5       \\ \hline
		\end{tabular}
	}
	\caption{Part localization results (\%) on CUB200-2011 with different numbers of nearest neighbors ($M$) and different bounding box overlapping thresholds.}
	\label{M_score}
\end{table}

\begin{table}[t!]
	\centering
	\resizebox{1\hsize}{!}{%
		\begin{tabular}{|l|c|c|c|c|}
			\hline
			\multicolumn{1}{|c|}{\multirow{2}{*}{Methods}} & \multicolumn{2}{c|}{Oracle Box Given} & \multicolumn{2}{l|}{Oracle Box Unknown} \\ \cline{2-5}
			\multicolumn{1}{|c|}{}                         & Head &              Body              & Head &               Body               \\ \hline
			Strong DPM~\cite{Azizpour2012Object}           & 43.5 &              75.2              & 37.4 &               47.1               \\ \hline
			Part-based R-CNNs~\cite{Zhang2014Part}         & \textbf{68.5} &              79.8              & \textbf{61.9} &               70.7               \\ \hline
			Transfer ($\geq0.5$)                           & 52.7 &              \textbf{90.6}              & 50.1 &               \textbf{85.9}               \\ \hline \hline
			Transfer ($\geq0.4$)                           & 70.2 &              95.3              & 68.0 &               92.1               \\ \hline
			Transfer ($\geq0.3$)                           & 80.7 &              98.2              & 80.5 &               96.0               \\ \hline
		\end{tabular}
	}
	\vspace{-0.1in}
	\caption{Comparison of part localization results (\%) on CUB200-2011. We also report the results with given oracle (ground-truth) object bounding boxes.}
	\label{part-location}
\end{table}

\begin{table}[t!]
	\begin{center}
		\begin{tabular}{|l|c|}
			\hline
			Input Image Region                     & Accuracy (\%) \\ \hline
			Entire Image                           &     62.5      \\ \hline
			Object-level Box (Oracle)              &     79.1      \\ \hline
			Object-level Box (Ours)                &     76.9      \\ \hline
			Head Box (Ours)                        &     67.4      \\ \hline
			Body Box (Ours)                        &     74.0      \\ \hline
		\end{tabular}
	\end{center}
	\vspace{-0.15in}
	\caption{Recognition results on CUB200-2011 using features computed from different object/part bounding boxes.}
	\label{part-result}
	\vspace{-0.15in}
\end{table}

\begin{table}[t!]
	\centering
	\resizebox{0.48\textwidth}{!}{%
		\begin{tabular}{|l|c|c|c|c|c|c|}
			\hline
			\multirow{2}{*}{Method}                                        & \multicolumn{2}{c|}{Train (Oracle)} & \multicolumn{2}{c|}{Test (Oracle)} & \multirow{2}{*}{Feature}  & \multirow{2}{*}{Accuracy (\%)}   \\ \cline{2-5}
			& Object &         Parts         & Object &        Parts         &                                                           &  \\ \hline
			Berg \em{et al.}~\cite{berg-poof-cvpr2013} (CVPR13)                                & ${\surd}$  &         ${\surd}$         & ${\surd}$  &        ${\surd}$         &           POOF            &              73.3                \\ \hline
			Zhang \em{et al.}~\cite{Zhang2014Part} (ECCV14)                           & ${\surd}$  &         ${\surd}$         & ${\surd}$  &        ${\surd}$         &          AlexNet          &              82.0                \\ \hline
			Branson \em{et al.}~\cite{Branson2014Bird} (BMVC14)      & ${\surd}$  &         ${\surd}$         & ${\surd}$  &        ${\surd}$         &            AlexNet             & \textbf{85.4} \\ \hline
			Goring \em{et al.}~\cite{goering2014nonparametric} (CVPR14)   & ${\surd}$  &         ${\surd}$         & ${\surd}$  &                          &            HOG            &              57.8                \\ \hline
			Gavves \em{et al.}~\cite{Gavves2013Fine} (ICCV13)                               & ${\surd}$  &         ${\surd}$         & ${\surd}$  &                          &          Fisher           &              62.7                \\ \hline
			Huang \em{et al.}~\cite{huang2016part} (CVPR16)                          & ${\surd}$  &         ${\surd}$         &       ${\surd}$     &                          &          AlexNet          &              \textbf{76.6}                \\ \hline
			Zhang \em{et al.}~\cite{Zhang2014Part} (ECCV14)                          & ${\surd}$  &         ${\surd}$         &            &                          &          AlexNet          &              73.9                \\ \hline
			Branson \em{et al.}~\cite{Branson2014Bird} (BMVC14)                        & ${\surd}$  &         ${\surd}$         &            &                          &          AlexNet          &              75.7                \\ \hline
			Zhang \em{et al.}$^\dag$~\cite{wang2015multiple} (ICCV15)         & ${\surd}$  &         ${\surd}$         &            &                          &          VGGNet           &              81.6                \\ \hline
			\textbf{Ours}                      & ${\surd}$  &         ${\surd}$         &            &                          &          VGGNet           &         \textbf{84.0}            \\ \hline
		\end{tabular}
	}
	\caption{Comparison of recognition results with the state of the art on CUB200-2011, organized based on the amount of used bounding box annotations in \emph{testing} images. The result of Zhang et al.$^\dag$ is from~\cite{wang2015multiple}, where the authors re-implemented Part-based RCNN~\cite{Zhang2014Part} with VGGNet.} 
	\label{CUB-2011results}
	\vspace{-0.15in}
\end{table}

\subsubsection{Recognition} 
We first discuss results using features computed from different object/part bounding boxes, in order to understand the contribution of each image region in fine-grained recognition. As shown in Table~\ref{part-result}, the accuracy of using features computed on the entire images (without localization) is worse than that relies on the features in the small head boxes (62.5\% and 67.4\% respectively). This indicates that using entire images is not reliable due to background clutter. Based on features computed in our predicted object or body bounding boxes, we achieve much better results. 

Table~\ref{CUB-2011results} gives the result from fusing the features computed in our predicted bounding boxes and the entire images (the bottom row), and compares it with a large set of approaches proposed recently. Fusing the features offers a big leap in the recognition performance, which validates the fact that it is important to focus on both the object and its important parts for fine-grained recognition. The compared approaches are grouped into three categories: the first three adopted additionally the ground-truth object and part bounding boxes in the test set; the next three used the ground-truth object bounding boxes in the test set; and the following three using neither object nor part annotations in the test set. Our approach, which performs automatic localization of objects and parts at test time, offers very competitive results.

\begin{table}
	\begin{center}
		\begin{tabular}{|l|c|}
			\hline
			Method                                                & Accuracy (\%) \\ \hline
			One vs. Most + ST Prior~\cite{berg-birdsnap-cvpr2014} &     66.6      \\ \hline\hline
			Entire Image Classification                           &     60.7      \\ \hline
			Object-level Box (Ours)                               & \textbf{73.4} \\ \hline
		\end{tabular}
	\end{center}
	\vspace{-0.15in}
	\caption{Recognition results on Birdsnap.}
	\vspace{-0.1in}
	\label{Birdsnap}
\end{table}

\subsection{Results on Birdsnap}
Finally, we present results on the Birdsnap dataset in Table~\ref{Birdsnap}. We see that the recognition accuracy of using the entire images is just 60.7\%, which is much lower than the result from the owner of the dataset~\cite{berg-birdsnap-cvpr2014}. By adopting our iterative object localization and using features from the predicted boxes, the performance can be significantly improved to 73.4\%, which again verifies the effectiveness of our approach. Notice that this large dataset does not contain part-level bounding box annotations and therefore the part-level transfer is not evaluated. 

\section{Conclusions}
We have proposed a novel approach for object and part localization in fine-grained recognition tasks. Our approach follows a data-driven pipeline by iteratively transferring bounding boxes from similar training images. 
We show that such a simple approach can produce better localization results than the popular proposal-based methods that have to filter thousands of candidate bounding box proposals in each image. Using deep learning features computed in our predicted object/part bounding boxes, very competitive accuracies are obtained. The results indicate that it is very important to incorporate clues from objects and parts so that the subtle differences across the categories can be captured. 

\noindent{\textbf{Acknowledgements}}
This work was supported in part by two NSFC projects (\#61572134 and \#U1509206) and one grant from STCSM, China (\#16QA1400500).

\footnotesize
\bibliographystyle{IEEEbib}
\bibliography{camera-ready_icme2017template}

\begin{thebibliography}{10}

\bibitem{Dalal2005Histograms}
Navneet Dalal and Bill Triggs,
\newblock ``Histograms of oriented gradients for human detection,''
\newblock in {\em CVPR}, 2005.

\bibitem{Zhang2014Part}
Ning Zhang, Jeff Donahue, and et~al.,
\newblock ``Part-based r-cnns for fine-grained category detection,''
\newblock in {\em ECCV}, 2014.

\bibitem{berg-poof-cvpr2013}
Thomas Berg and Peter~N Belhumeur,
\newblock ``{POOF}: {P}art-{B}ased {O}ne-vs-{O}ne {F}eatures for fine-grained
  categorization, face verification, and attribute estimation,''
\newblock in {\em CVPR}, 2013.

\bibitem{Branson2014Bird}
Steve Branson, Grant Van~Horn, and et~al.,
\newblock ``Bird species categorization using pose normalized deep
  convolutional nets,''
\newblock in {\em BMVC}, 2014.

\bibitem{Zhang2013Deformable}
Ning Zhang, Ryan Farrell, and et~al.,
\newblock ``Deformable part descriptors for fine-grained recognition and
  attribute prediction,''
\newblock in {\em ICCV}, 2013.

\bibitem{felzenszwalb2010object}
Pedro~F Felzenszwalb, Ross~B Girshick, David McAllester, and Deva Ramanan,
\newblock ``Object detection with discriminatively trained part-based models,''
\newblock {\em PAMI}, vol. 32, no. 9, pp. 1627--1645, 2010.

\bibitem{torralba200880}
Antonio Torralba, Rob Fergus, and William~T Freeman,
\newblock ``80 million tiny images: A large data set for nonparametric object
  and scene recognition,''
\newblock {\em PAMI}, vol. 30, no. 11, pp. 1958--1970, 2008.

\bibitem{ren2005data}
Liu Ren, Alton Patrick, Alexei~A Efros, Jessica~K Hodgins, and James~M Rehg,
\newblock ``A data-driven approach to quantifying natural human motion,''
\newblock {\em ACM TOG}, vol. 24, no. 3, pp. 1090--1097, 2005.

\bibitem{goering2014nonparametric}
Christoph Goring, Erid Rodner, and et~al.,
\newblock ``Nonparametric part transfer for fine-grained recognition,''
\newblock in {\em CVPR}, 2014.

\bibitem{huang2016part}
Shaoli Huang, Zhe Xu, Dacheng Tao, and Ya~Zhang,
\newblock ``Part-stacked cnn for fine-grained visual categorization,''
\newblock in {\em CVPR}, 2016.

\bibitem{pu2014looks}
Jian Pu, Yu-Gang Jiang, Jun Wang, and Xiangyang Xue,
\newblock ``Which looks like which: Exploring inter-class relationships in
  fine-grained visual categorization,''
\newblock in {\em ECCV}, 2014.

\bibitem{Kumar2012Leafsnap}
Neeraj Kumar, Peter~N Belhumeur, and et~al.,
\newblock ``Leafsnap: A computer vision system for automatic plant species
  identification,''
\newblock in {\em ECCV}, 2012.

\bibitem{cui2016fine}
Yin Cui, Feng Zhou, Yuanqing Lin, and Serge Belongie,
\newblock ``Fine-grained categorization and dataset bootstrapping using deep
  metric learning with humans in the loop,''
\newblock in {\em CVPR}, 2016.

\bibitem{Aditya2011}
Aditya Khosla, Nityananda Jayadevaprakash, Bangpeng Yao, and Fei-Fei Li,
\newblock ``Novel dataset for fine-grained image categoraization: Stanford
  dogs,''
\newblock in {\em CVPR Workshop on FGVC}, 2011.

\bibitem{xiao2014application}
Tianjun Xiao, Yichong Xu, and et~al.,
\newblock ``The application of two-level attention models in deep convolutional
  neural network for fine-grained image classification,''
\newblock in {\em CVPR}, 2015.

\bibitem{Wah2011Multiclass}
Catherine Wah, Steve Branson, Pietro Perona, and Serge Belongie,
\newblock ``Multiclass recognition and part localization with humans in the
  loop,''
\newblock in {\em ICCV}, 2011.

\bibitem{Branson2011Strong}
Steve Branson, Pietro Perona, and et~al.,
\newblock ``Strong supervision from weak annotation: Interactive training of
  deformable part models,''
\newblock in {\em ICCV}, 2011.

\bibitem{carreira2015human}
Joao Carreira, Pulkit Agrawal, and et~al.,
\newblock ``Human pose estimation with iterative error feedback,''
\newblock {\em arXiv preprint arXiv:1507.06550}, 2015.

\bibitem{simonyan2014very}
Karen Simonyan and Andrew Zisserman,
\newblock ``Very deep convolutional networks for large-scale image
  recognition,''
\newblock {\em arXiv preprint arXiv:1409.1556}, 2014.

\bibitem{he2014spatial}
Kaiming He, Xiangyu Zhang, Shaoqing Ren, and Jian Sun,
\newblock ``Spatial pyramid pooling in deep convolutional networks for visual
  recognition,''
\newblock in {\em ECCV}, 2014.

\bibitem{R.2011Birdlets}
Ryan Farrell, Om~Oza, Ning Zhang, Vlad~I Morariu, Trevor Darrell, and Larry~S
  Davis,
\newblock ``Birdlets: Subordinate categorization using volumetric primitives
  and pose-normalized appearance,''
\newblock in {\em ICCV}, 2011.

\bibitem{Liu2012Dog}
Jiongxin Liu, Angjoo Kanazawa, David Jacobs, and Peter Belhumeur,
\newblock ``Dog breed classification using part localization,''
\newblock in {\em ECCV}, 2012.

\bibitem{Girshick2013Rich}
Ross Girshick, Jeff Donahue, Trevor Darrell, and Jitendra Malik,
\newblock ``Rich feature hierarchies for accurate object detection and semantic
  segmentation,''
\newblock in {\em CVPR}, 2014.

\bibitem{mm14:videoclassification}
Z.~Wu, Y.-G. Jiang, J.~Wang, J.~Pu, and X.~Xue,
\newblock ``Exploring inter-feature and inter-class relationships with deep
  neural networks for video classification,''
\newblock in {\em ACM MM}, 2014.

\bibitem{Wah2011The}
Catherine Wah, Steve Branson, Peter Welinder, Pietro Perona, and Serge
  Belongie,
\newblock ``The caltech-ucsd birds-200-2011 dataset,''
\newblock {\em California Institute of Technology}, 2011.

\bibitem{berg-birdsnap-cvpr2014}
Thomas Berg, Jiongxin Liu, Seung~Woo Lee, Michelle~L. Alexander, David~W.
  Jacobs, and Peter~N. Belhumeur,
\newblock ``Birdsnap: Large-scale fine-grained visual categorization of
  birds,''
\newblock in {\em CVPR}, 2014.

\bibitem{Azizpour2012Object}
Hossein Azizpour and Ivan Laptev,
\newblock ``Object detection using strongly-supervised deformable part
  models,''
\newblock in {\em ECCV}, 2012.

\bibitem{Gavves2013Fine}
Efstratios Gavves, Basura Fernando, and et~al.,
\newblock ``Fine-grained categorization by alignments,''
\newblock in {\em ICCV}, 2013.

\bibitem{wang2015multiple}
Dequan Wang, Zhiqiang Shen, Jie Shao, Wei Zhang, Xiangyang Xue, and Zheng
  Zhang,
\newblock ``Multiple granularity descriptors for fine-grained categorization,''
\newblock in {\em ICCV}, 2015.

\end{thebibliography}

\end{document}